\documentclass[letterpaper]{article} 
\usepackage{aaai24}  
\usepackage{times}  
\usepackage{helvet}  
\usepackage{courier}  
\usepackage[hyphens]{url}  
\usepackage{graphicx} 
\usepackage{natbib}  
\usepackage{caption} 
\usepackage{algorithm}
\usepackage{algorithmic}
\usepackage{amsmath}
\usepackage{amssymb}
\usepackage{multirow}
\usepackage{diagbox}
\usepackage{colortbl}
\usepackage{booktabs}
\usepackage{bbding}
\usepackage{makecell}
\usepackage{csquotes}
\usepackage{tikz}

\newif\ifstartedinmathmode
\newcommand\encircled[1]{%
	\relax\ifmmode\startedinmathmodetrue\else\startedinmathmodefalse\fi%
	\tikz[baseline,anchor=base]{%
		\node[draw,circle,outer sep=0pt,inner sep=.2ex]
		{\ifstartedinmathmode$#1$\else#1\fi};}%
}

\usepackage{newfloat}
\usepackage{listings}
\DeclareCaptionStyle{ruled}{labelfont=normalfont,labelsep=colon,strut=off} 
\floatstyle{ruled}
\newfloat{listing}{tb}{lst}{}
\floatname{listing}{Listing}
\title{UniST: Towards Unifying Saliency Transformer for Video Saliency Prediction and Detection}

\author{
	Junwen Xiong\textsuperscript{\rm 1}\equalcontrib, 
	Peng Zhang\textsuperscript{\rm 1,\rm 2}\equalcontrib \thanks{ Corresponding author.}
	Chuanyue Li\textsuperscript{\rm 1}\equalcontrib, \\
	Wei Huang\textsuperscript{\rm 3}, 
	Yufei Zha\textsuperscript{\rm 1,\rm 2},
	Tao You\textsuperscript{\rm 1}
}
\affiliations{
    \textsuperscript{\rm 1}Northwestern Polytechnical University, \textsuperscript{\rm 2}Ningbo Institute of Northwestern Polytechnical University  \\
    \textsuperscript{\rm 3}Nanchang University\\
    wshzxjw@mail.nwpu.edu.cn, zh0036ng@nwpu.edu.cn

}

\usepackage{bibentry}

\begin{document}

\maketitle

\begin{abstract}
Video saliency prediction and detection are thriving research domains that enable computers to simulate the distribution of visual attention akin to how humans perceiving dynamic scenes.
While many approaches have crafted task-specific training paradigms for either video saliency prediction or video salient object detection tasks, few attention has been devoted to devising a generalized saliency modeling framework that seamlessly bridges both these distinct tasks.
In this study, we introduce the \textbf{Uni}fied \textbf{S}aliency \textbf{T}ransformer (\textbf{UniST}) framework, which  comprehensively utilizes the essential attributes of video saliency prediction and video salient object detection. 
In addition to extracting representations of frame sequences, a saliency-aware transformer is designed to learn the spatio-temporal representations at progressively  increased resolutions, while incorporating effective cross-scale  saliency information to produce a robust representation.  
Furthermore, a task-specific decoder is proposed to perform the final prediction for each task.  To the best of our knowledge, this is the first work that explores designing a transformer structure for both saliency modeling tasks. Convincible experiments demonstrate that the proposed \textbf{UniST} achieves superior performance across seven challenging  benchmarks for two tasks, and significantly outperforms the other state-of-the-art methods. 

\end{abstract}

\section{Introduction}

As the continuous emergence of massive dynamic data propels deep learning further towards human vision, a growing attention of research has focused on video saliency prediction (VSP) \cite{bak2018spatio, wang2018revisiting, tsiami2020stavis} and video salient object detection (VSOD) \cite{mei2021transvos, zhang2021dynamic, liu2022ds}, which are regarded as the fundamental tasks in computer vision. Both the tasks essentially aim to simulate the visual attention distribution of humans perceiving dynamic scenes by modeling spatio-temporal cues in video content. However, the modeling paradigms of prior works are tailored for specialized tasks, and thereby lack the capacity for generalization to address broader tasks.




\begin{figure}[t]
	\includegraphics[scale=0.5]{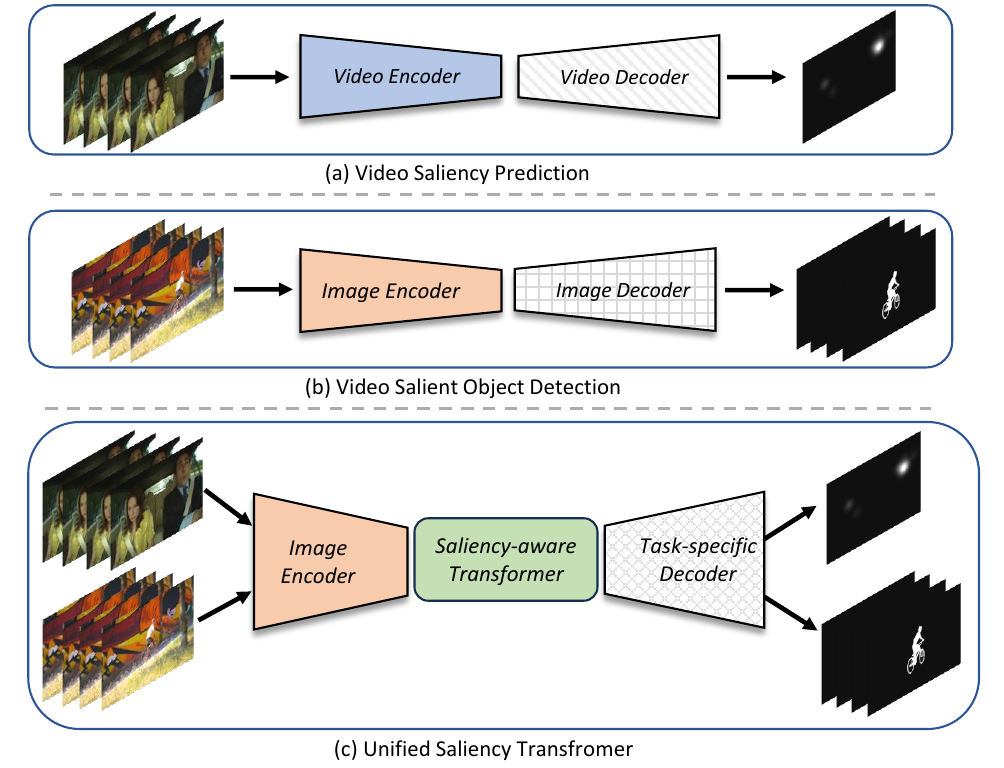}
	\caption{Comparisons of the traditional different modeling paradigm for VSP and VSOD tasks, as well as our proposed unified saliency transformer framework. The VSP and VSOD adopt video encoder and image encoder respectively as feature extractors, followed by corresponding decoders. Differently, a unified saliency transformer directly applies an image encoder for feature processing, and follows a transformer structure for spatio-temporal modeling, and finally uses different decoders for different tasks. }
	\label{fig-avi}
\end{figure}



Previous VSP approaches have made impressive progress in predicting the most salient visual regions within frame sequences based on video encoder and decoder \cite{tsiami2020stavis, jain2021vinet,  zhou2023transformer, xiong2023casp}, as depicted in Figure \ref{fig-avi}(a). To capture spatio-temporal information from video sequences, these methods uniformly rely on video models pre-trained on extensive video datasets, e.g., the lightweight \textit{S3D}  \cite{xie2018rethinking} and the effective \textit{Video Swin Transformer} \cite{liu2022video}. Subsequently, various kinds of decoding strategies have been successively proposed for processing the obtained spatio-temporal features. \cite{min2019tased, jain2021vinet} proposed a 3D fully convolutional decoder with U-Net-like structure to progressively concatenate features along the temporal dimension. In addition, \cite{zhou2023transformer} proposed a 3D convolutional decoder based on the feature pyramid structure, to build high-level semantic features at all scales.


%

Compared to VSP, the video dataset used in the current VSOD task is usually limited-sized, making it difficult to converge the video encoder-based method to the optimum \cite{le2017deeply, fan2019shifting}. As a result, many VSOD approaches adopt an image encoder and decoder-based training paradigm, where pre-training is initially performed on the image dataset, and then the pre-trained weights are transferred to the video dataset to further improve the model's generalization capabilities \cite{gu2020pyramid,  zhang2021dynamic, liu2022ds}, as shown in the Figure \ref{fig-avi}(b). To extract the frame-wise features from video sequences, \cite{gu2020pyramid, zhang2021dynamic, liu2022ds} employed networks of the ResNet family \cite{he2016deep} and the MobileNet family \cite{howard2017mobilenets} as feature extractors. For the temporal cues between image feature sequences, \cite{gu2020pyramid} proposed to use the modified non-local blocks \cite{wang2018non} to obtain the temporal information between feature sequences. For temporal features extraction in a different way, \cite{zhang2021dynamic, liu2022ds} proposed to construct another temporal branch from optical flow and fuse with spatial features.


Unfortunately, except for the above achievements on each independent saliency task, there has not been much prior effort to bridge both tasks for the construction of a generalized saliency modeling paradigm. Thus, several questions naturally arise: 1) \textit{why it is difficult to unify modeling for video saliency prediction and detection tasks?} and 2) \textit{is it possible to build a unified saliency model generalized to these two different tasks?}

As an answer to the questions above, a novel \textbf{Uni}fed \textbf{S}aliency \textbf{T}ransformer model (\textbf{UniST}) is proposed, which  comprehensively utilizes the essential attributes of video saliency prediction and video salient object detection tasks. \textbf{UniST} composes of an image encoder, a saliency-aware transformer and a task-specific decoder as shown in Figure \ref{fig-avi}(c), and the incorporated image encoder is to obtain a generic representation for each image with video sequences.
In addition, a saliency-aware transformer is also introduced to model spatio-temporal representations of image feature sequences by stacking multiple sal-transformer blocks, as well as augmenting feature scale progressively. Subsequently, a task-specific decoder is proposed to leverage the transformer's output to make the final prediction for each task. We train \textbf{UniST} and achieve superior performance to well-established prior works for both tasks. The main contributions in this work can be summarized as follows:


\begin{itemize}
	\item [1)] A novel unified saliency transformer \textbf{UniST} framework is proposed to comprehensively intrinsically associate both video saliency prediction and video salient object detection tasks.
	\item[2)] An efficient saliency-aware transformer is designed to learn the spatio-temporal representations at gradually increased resolutions, and incorporate effective cross-scale saliency information in the meantime.
	\item[3)] Convincible experiments have been conducted on different challenging benchmarks across two tasks, which is able to demonstrate a superior performance of the proposed \textbf{UniST} in comparison to the other state-of-the-art works.
	
\end{itemize}

\section{Related Work}

\subsection{Video Saliency Prediction}

Deep learning has led to the emergence of numerous video saliency prediction methods that focus on modeling continuous motion information across frames.  \cite{bak2017spatio} presented a two-stream deep model based on element-wise or convolutional fusion strategies to learn spatio-temporal features. \cite{jiang2017predicting} designed an object-to-motion CNN network to extract features and two-layer ConvLSTM with a Bayesian dropout to learn dynamic saliency. \cite{wang2018revisiting} combined the ConvLSTM network with an attention mechanism to improve training efficiency and performance. However,  existing LSTM-based methods fall short of effectively integrating spatial and temporal information \cite{min2019tased}. For this challenge, different studies have tried to incorporate 3D convolutional models or video transformer models into the VSP task. \cite{min2019tased} introduced a TASED-Net to leverage the S3D model \cite{xie2018rethinking} to simultaneously handle spatial and temporal cues in VSP. Since then, the training paradigm of 3D convolution has been widely used in VSP task. \cite{jain2021vinet} adopted a 3D encoder-decoder structure, resembling U-Net, which allows the constant concatenation of decoding features from various layers(with the corresponding encoder features in the temporal dimension). \cite{ma2022video} proposed a video swin transformer-based framework, which can eliminate the reliance on pre-trained S3D model and enhance the performance upper bound in VSP task. In a different way, the proposed UniST directly applies an image encoder for feature processing and follows a saliency-aware transformer for spatio-temporal modeling.

\begin{figure*}[!htbp]
	\centering
	\includegraphics[width=0.9\textwidth]{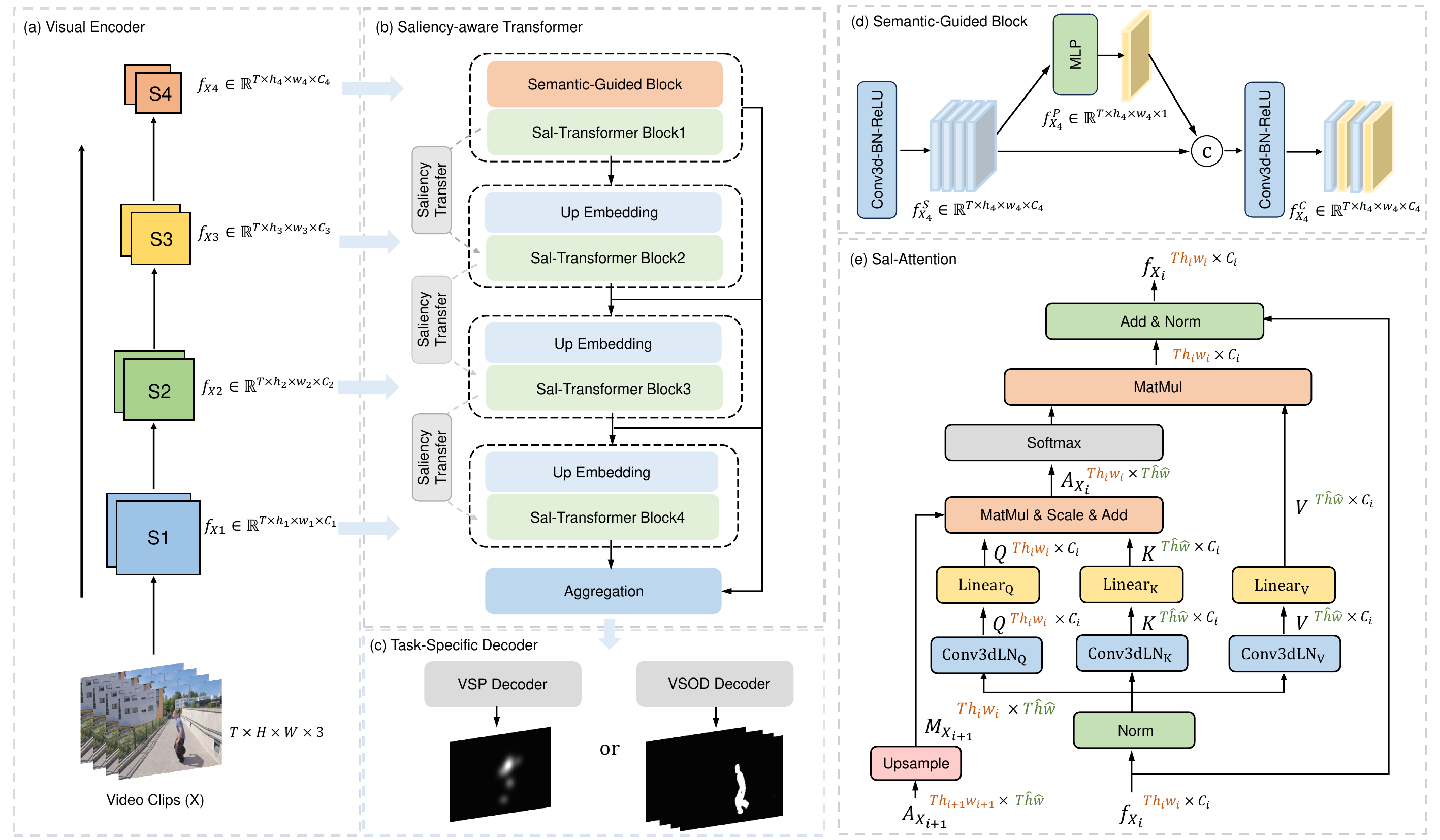}
	
	\caption{An overview of the proposed UniST. The visual encoder learns frame-wise visual representations from the video clips. The multi-scale visual features are then used as inputs to the saliency-aware transformer for global spatio-temporal modeling, generating refined and scaled-up spatio-temporal features for final prediction. \encircled{c} denotes the channel-wise concatenation.
	}
	\label{fig-model_structure}
	
\end{figure*}

\subsection{Video Salient Object Detection}



In VSOD, since the content of each frame is highly correlated, it can be considered whose purpose is to capture long-range feature information among the adjacency frame. Traditional methods \cite{girshick2014rich, liu2016dhsnet} often rely on conventional heuristics drawn from the domain of image salient object detection. Recent works \cite{le2017deeply, wang2017video} strive to acquire highly semantic representations and usually perform spatial-temporal detection end-to-end.

By taking temporal information into consideration networks modeling, different works have been proposed, e.g., ConvLSTM \cite{li2018flow}, took optical-flows as input \cite{li2019motion}, or 3D convolution \cite{miao2020memory}. But in real-world scenarios, the temporal cost incurred by introducing three-dimensional convolution operations is noteworthy, and incorporating optical flow information may not fully align with the ideal concept of an end-to-end network. More recently, attention-based mechanisms have gained traction for refining pairwise relationships between regions across consecutive frames. \cite{fan2019shifting, gu2020pyramid} proposed a visual-attention-consistent module and a pyramid constrained self-attention block to better capture the temporal dynamics cues, respectively. Despite these advancements, the existing methods remain tailored to specific tasks. Therefore,  this paper proposes a unified framework to address VSOD and VSP tasks more comprehensively.

\section{Proposed Method}


An overview of the proposed UniST is presented in Figure \ref{fig-model_structure}. To tackle the challenges of video saliency prediction and video salient object detection, the UniST is constructed based on the encoder-decoder architecture which consists of a visual feature encoder, a saliency-aware transformer and a task-specific decoder.

To elaborate, video clips are initially fed into the visual feature encoder, yielding multi-level spatial features for each image. Then, a saliency-aware transformer is introduced to intricately capture spatio-temporal representations of image feature sequences. This is achieved through the stacking of multiple effective sal-transformer stages, which increases the scale of the feature maps progressively. Finally, a task-specific decoder is devised to leverage the transformer's output features, and facilitate the output predictions for each individual task.

\subsection{Visual Feature Encoder}

Let $X \in  \mathbb{R}^{   T \times H  \times W \times  3 }$ denote an RGB video clip of length $T$. This is input to a 2D backbone network which produces frame-wise feature maps. The backbone consists of 4 encoder stages, and outputs 4 hierarchical visual feature maps, illustrated in Figure \ref{fig-model_structure}(a). The generated feature maps are denoted as $\{ f_{X_i} \}  \in \mathbb{R}^{ T \times h_i  \times w_i \times C_i} $, where $(h_i, w_i) = (H, W)/2^{i+1}, i=1,...,4$. In practical implementation, we employ the off-the-shelf MViT\cite{fan2021multiscale} as a visual encoder to encode the spatial information of image sequences, which can also be replaced with other general-purpose encoders, e.g., PVT\cite{wang2021pyramid}, Swin\cite{liu2021swin}.

\begin{equation}
	\begin{aligned}
		&\{f_{X_1}, f_{X_2}, f_{X_3}, f_{X_4} \} = VisualEncoder(X) \\
	\end{aligned}
\end{equation}

\subsection{Saliency-aware Transformer}
For each input video clip, the encoder generates feature maps at four distinct scales. Nevertheless, the extracted multi-scale feature maps solely capture information within the spatial domain of the image sequences, neglecting any temporal domain modeling. Another issue to be noted is that the low spatial resolution of  the visual encoder's output feature makes it unsuitable for VSP and VSOD tasks.
With these considerations, a saliency-aware transformer is designed to conduct spatio-temporal modeling, as well as enhance the resolution of the feature maps progressively.

As shown in Figure \ref{fig-model_structure}(b), there are four stages in the saliency-aware transformer, and each one is a designed sal-transformer stage.
The primary  stage of the saliency-aware transformer focus on learning the spatio-temporal sal-attention of the feature with the lowest resolution, and the subsequent three stages augment the spatial resolution of the feature maps, and calculate spatio-temporal sal-attention at higher resolutions. Similarly, the semantic-guided block and sal-transformer block are used together in the first stage, but the up embedding block and sal-transformer block are used in the following three stages.


\subsubsection{Semantic-Guided Block}


Considering the most substantial semantic cues encapsulated by high-level features\cite{chang2021temporal, sun2022hierarchical}, a semantic-guided block is introduced to effectively guide the saliency modeling process.
Specifically, as in Figure \ref{fig-model_structure}(d), $f_{X_4}$ is initially fed into a Conv3d-BN-ReLU block denoted as $F_S$. It consists of a 3D convolutional layer with kernel size of $3 \times 3 \times 3$, a 3D batch nomalization layer and a ReLU activation function. The output of this block is a  semantic feature $f_{X_4}^S$. This semantic feature is processed by a linear projection layer $F_P$ to reduce the channel dimension to 1, and generate a saliency feature  map $f_{X_4}^{P}$. 
Finally, the semantic and saliency feature maps are concatenated along the channel dimension,  and we use another Conv3d-BN-ReLU block $F_C$ to adjust the channel numbers to the original dimension $C_4$ to obtain the combined feature $f_{X_4}^{C} \in \mathbb{R}^{ T \times h_4  \times w_4 \times C_4 }$.

\begin{equation}
	\begin{aligned}
		&f_{X_4}^{C} = F_{C}(Cat(F_{P}(F_{S}(f_{X_4})), F_{S}(f_{X_4}))) \\
	\end{aligned}
\end{equation}

\noindent where $Cat(\cdot, \cdot)$ represents the concatenation operation. To feed the sementic-guided feature into the subsequent sal-transformer block,  we flatten the feature in the spatio-temporal dimension to obtain a 2D feature-token sequence $f_{X_4}^{C} \in \mathbb{R}^{ Th_4 w_4 \times C_4 }$.

\subsubsection{Up Embedding}


Since the transformer module typically operates on 2D feature-token sequences, which may cause the original spatial layout of the feature map has been corrupted.
To reconstruct the spatial details within  the feature map, an up embedding block specifically has been designed for the sal-transformer block.  The input to each up embedding block comes from  the 2D feature-token sequence, which is generated by the previous sal-transformer stage. 
In the beginning, the up embedding block reshapes the feature sequence  $f_{X_i} \in \mathbb{R}^{Th_iw_i \times C_i}$ into spatial-temporal feature maps of dimensions  $\mathbb{R}^{T \times h_i \times w_i \times C_i}$. 
Subsequently, a bilinear interpolation is performed to amplify the height and width of each spatial feature by $2\times$ along the temporal dimension, and a Conv-BN-ReLU block $F_{Conv}$ is used to reduce the channel dimension to $C_{i-1}$ at the same time. For the obtained features of size $\mathbb{R}^{T \times h_{i-1} \times w_{i-1} \times C_{i-1}}$, they are fused with the feature maps of previous hierarchical level $f_{X_{i-1}}$ via addition operation.



\begin{equation}
	\begin{aligned}
		&f_{X_{i-1}} =  f_{X_{i-1}} + F_{Conv}(Interpolation(f_{X_i}))  \\
	\end{aligned}
\end{equation}

The fused features are reshaped back to feature sequence  $f_{X_{i-1}} \in \mathbb{R}^{Th_{i-1}w_{i-1} \times C_{i-1}}$ as an upsampled token sequence.

\subsubsection{Sal-Transformer Block}
When the feature maps obtained by the semantic-guided or up embedding block are input into the sal-transformer block, spatio-temporal feature modeling starts. Inspired by incorporating multi-scale feature maps to improve model performance \cite{li2019motion,jain2021vinet}, sal-transformer block is proposed which not only models the spatio-temporal domain of features via sal-attention mechanism, but also integrates the multi-scale attention information of the previous stage.



Notice that calculating global sal-attention from temporal high-resolution features has a prohibitively large memory footprint. To alleviate this problem, in sal-attention (see shown in Figure \ref{fig-model_structure} (e)), 
a reduction operation is performed on the size of query $Q$, key $K$, and value $V$ matrices for attention computation, like prior works \cite{fan2021multiscale, wang2021pyramid}.
In detail, it utilizes the  $Conv3dLN$ operation consisting of a 3D convolution and a layer normalization for embedding extraction, and obtains $Q$, $K$ and $V$ in varying dimensions by controlling the convolution's kernel and stride sizes.
This strategy for dimensionality reduction  considerably enhances the memory and computational efficiency associated with global sal-attention computation, which makes it feasible to be used by multiple consecutive sal-transformer blocks.

For an input feature $ f_{X_i} \in \mathbb{R}^{Th_i w_i \times C_i}$, the sal-attention first transforms the 2D feature sequence into spatio-temporal domain, and followed by the operation of $Conv3dLN_Q$, $Conv3dLN_K$ and $Conv3dLN_V$  to $f_{X_i}$. Then, the  linear projections  $W_Q, W_K, W_V \in \mathbb{R}^{C_i \times C_i}$ are applied to obtain $Q$, $K$ and $V$ embeddings, respectively.  The attention score matrix $A_{X_i}$ of $X_{i}$ feature map can be calculated as:

\begin{equation}
	\begin{aligned}
		&Q_{X_i} = W_Q(Conv3dLN_Q(f_{X_i})), Q_{X_i} \in \mathbb{R}^{Th_i w_i \times C_i} \\
		&K_{X_i} = W_K(Conv3dLN_K(f_{X_i})),  K_{X_i} \in \mathbb{R}^{T\hat{h} \hat{w} \times C_i} \\
		&V_{X_i} =  W_V(Conv3dLN_V(f_{X_i})),  V_{X_i} \in \mathbb{R}^{T\hat{h} \hat{w} \times C_i} \\
		&A_{X_i} = \frac{Q_{X_i}(K_{X_i})^T}{\sqrt{C_{i}}}, A_{X_i} \in \mathbb{R}^{Th_i w_i  \times T\hat{h} \hat{w} }
	\end{aligned}
\end{equation}


To further utilize the cross-scale insights from different stages within the saliency-aware transformer, a saliency transfer mechanism is introduced to enhance the attention score$A_{X_i}$ before  the softmax operation by integrating attention scores from distinct transformer stages.
 The attention score $A_{X_{i+1}}$ from the feature map $X_{i+1}$ is utilized to bolster the attention score $A_{X_i}$ of the feature map $X_i$.
It's noteworthy that the second dimension of attention scores across different stages maintains consistent dimensions because of the size of the convolutional kernel  designed.
Specifically, a reshape operation is performed to align the shape of $A_{X_{i+1}}$ with $\mathbb{R}^{T \times h_{i+1} \times w_{i+1} \times T\hat{h} \hat{w} }$, and followed by a $2\times$ bilinear interpolation in the first two spatial dimensions, then finally flatten it to get the attention matrix $M_{X_{i+1}} \in \mathbb{R}^{Th_i w_i  \times T\hat{h} \hat{w} }$  with same dimension as $A_{X_{i}}$.
The obtained $M_{X_{i+1}}$ is fused with  $A_{X_i}$ via addition to obtain the fused attention score with a convolution operation. Then, we perform a softmax operation on the score and a dot-product operation with $V$, the final updated feature maps $f_{X_i}$ becomes,

\begin{equation}
	\begin{aligned}
		&A_{X_i} = Conv(A_{X_i} + M_{X_{i+1}}) \\
		&f_{X_i} = Softmax(A_{X_i}) V_i + f_{X_i}
	\end{aligned}
\end{equation}

Since each sal-transformer stage produces feature maps with varying scales, an efficient strategy is also proposed to aggregate the multi-scale features. For each level of features, we employ a 3D convolution and upsample operation. The channel dimension of feature maps is initially adjusted by 3D convolution, and then their spatial dimensions are upsampled to align with the dimensions of the $f_{X_1}$ feature map. To obtain the aggregated feature $f_X \in \mathbb{R}^{T \times h_{1} \times w_{1} \times C_{1}}$, the features at each scale are concatenated along channel dimension and the number of channels is reduced by 3D convolution.


\subsection{Task-Specific Decoder}
\subsubsection{Video Saliency Prediction} For the feature $f_X$,  a combination of Conv3d-BN-ReLU block and Conv3d-Sigmoid block is employed. The former compresses the temporal dimension of the feature to 1, and the latter further reduces the channel dimension of the feature to 1. The upsampling operation is performed to output the prediction results $P_{VSP}$ by aligning the spatial dimension of ground-truth at the meantime.

\subsubsection{Video Salient Object Detection} As the VSOD task requires the output of $T$-frames, the temporal dimension compression of the feature $f_X$ is not necessary.  We directly apply a Conv-Sigmoid block to reduce the number of channels of $f_X$. Like the VSP decoder, the predictions $P_{VSOD}$ are upsampled to align the spatial dimensions of the ground-truth.

\section{Experiments}

\begin{table}[!tbp]
	\renewcommand\arraystretch{1.2}
	\resizebox{\linewidth}{!}{
		\begin{tabular}{lllll}
			\toprule
			\multirow{2}{*}{Method} & \multicolumn{2}{c}{\textbf{DHF1K}}  & \multicolumn{2}{c}{\textbf{$\mathit{\textrm{DAVIS}}_{16}$}}  \\
			\cmidrule(r){2-3}  \cmidrule(r){4-5}
			& $\mathit{CC}$ $\uparrow$  & $\mathit{SIM}$ $\uparrow$ & $\mathit{MAE}\downarrow$ & $S_m\uparrow$  \\
			\midrule\midrule
			UniST baseline                          & 0.521 & 0.410 & 0.025  & 0.880     \\
			\hline
			UniST w/SAT                              & 0.532 & 0.417 & 0.022  & 0.891    \\
			UniST w/SAT+SGB                          & 0.536 & 0.419 & 0.019  & 0.898    \\
			UniST w/SAT+SGB+ST   					& \textbf{0.541} & \textbf{0.423} & \textbf{0.018}  & \textbf{0.904}   \\
			\midrule
			Performance $\Delta$                    &+0.020 & +0.013 & -0.007  & +0.024     \\
			\bottomrule
	\end{tabular}}
	\caption{Ablation Studies. The proposed UniST and its components yield consistent improvement on different datasets and achieve clear overall improvement on both  tasks. $\downarrow$ means lower better and $\uparrow$ means higher better.}
	\label{table-ablation_1}
\end{table}

\subsection{Datasets}

For convincible validation, three popular video datasets, DHF1K\cite{wang2018revisiting}, Hollywood-2\cite{marszalek2009actions} and UCF-Sports\cite{rodriguez2008action}, are used in our experiment. DHF1K contains 600 training videos, 100 validation videos and 300 testing videos with a frame rate of 30 fps. The UniST model can only be evaluated on the validation set of DHF1K due to unavailable annotations of the test set. Hollywood2 contains 1707 videos extracted from 69 movies with 12 categorized action classes, 823 videos are used for training and 884 for testing. UCF-Sports contains 150 videos (103 for training, 47 for testing) collected from broadcast TV channels, which cover 9 sports, such as diving, weightlifting, and horse riding.

For VSOD, the proposed model is evaluated  on four public benchmark datasets including DAVIS$_{16}$\cite{perazzi2016benchmark}, FBMS\cite{ochs2013segmentation}, ViSal\cite{wang2015consistent} and SegTrackV2\cite{li2013video}. DAVIS$_{16}$ is a frequently used dataset, which contains 50 videos with a total of 3455 high-quality pixel-wise annotation frames. FBMS is a test dataset containing 59 videos with 720 sparsely annotated frames. ViSal is a dataset only used for test containing 19 videos with 193 pixel-wise annotation frames. SegTrackV2 is also a test dataset with 14 videos and 1,065 annotated frames.

\begin{figure}[!tbp]
	\flushright
	\begin{minipage}[t]{0.49\linewidth}
		\centering
		\includegraphics[width=\linewidth]{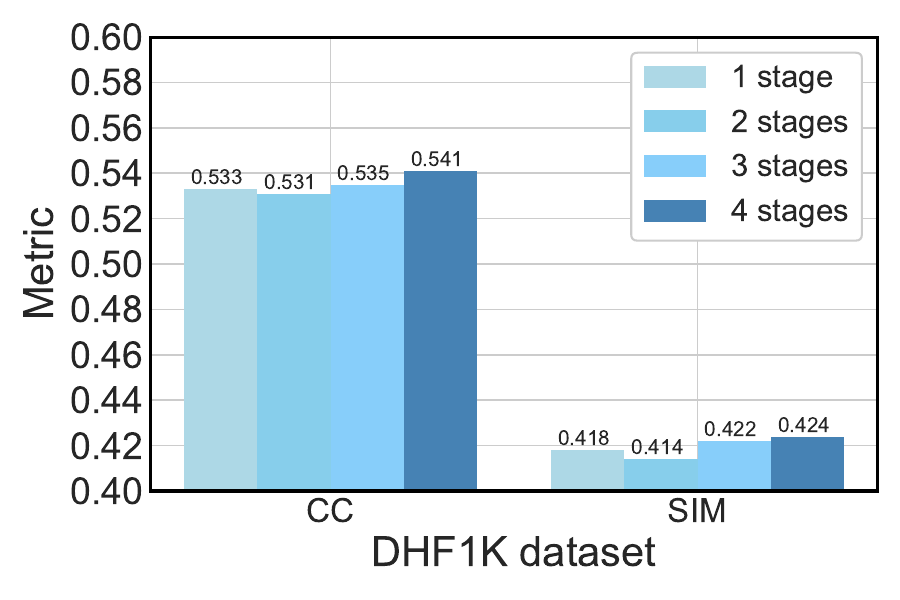}
	\end{minipage}
	\begin{minipage}[t]{0.49\linewidth}
		\centering
		\includegraphics[width=\linewidth]{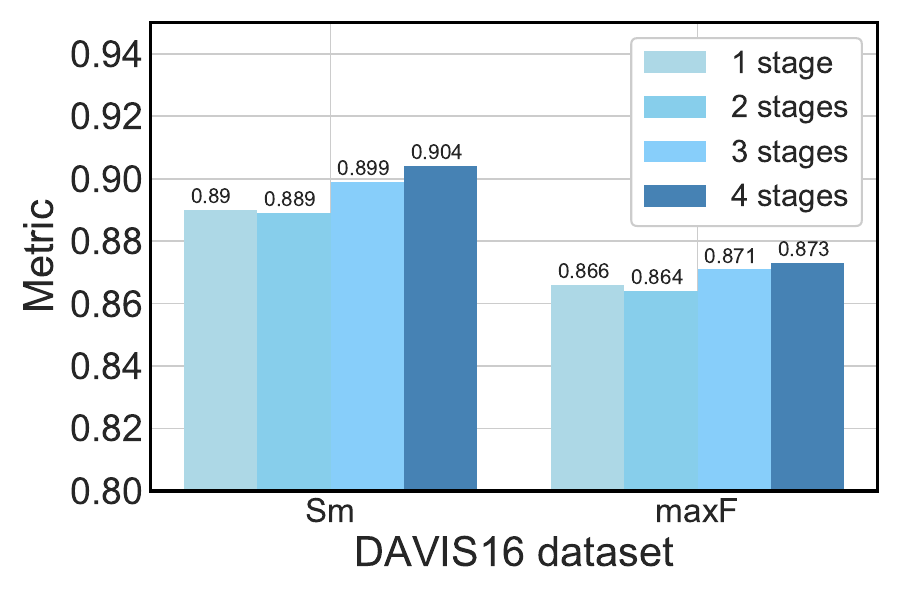}
	\end{minipage}
	\caption{Performance analysis of Sal-Transformer Stages on DHF1K and DAVIS$_{16}$ datasets.}
	\label{fig_transformer_stages}
\end{figure}

\begin{table}[!tbp]
	\centering
	\renewcommand\arraystretch{1.2}
	\resizebox{\linewidth}{!}{
		\begin{tabular}{lllllll}
			\toprule
			\multirow{2}{*}{Encoder} & \multicolumn{3}{c}{\textbf{DHF1K}}  & \multicolumn{3}{c}{\textbf{$\mathit{\textrm{DAVIS}}_{16}$}}  \\
			\cmidrule(r){2-4}  \cmidrule(r){5-7}
			& $\mathit{AUC}$-$J$ $\uparrow$ & $\mathit{CC}$ $\uparrow$  & $\mathit{SIM}$ $\uparrow$ & $\mathit{MAE}\downarrow$ & $S_m\uparrow$ & $F_\beta \uparrow$ \\
			\midrule\midrule
			Image MViT 	                        &0.920  & 0.541 & 0.423   & \textbf{0.018}  & \textbf{0.904} &\textbf{0.873}     \\
			Video MViT                            &\textbf{0.921} & \textbf{0.547} & \textbf{0.428}  & 0.069  & 0.729 & 0.590    \\
			\hline
			Image Swin							& 0.909 & 0.483 & 0.371 & 0.025  & 0.885  &0.854   \\
			Video Swin							& 0.911 & 0.510 & 0.384 & 0.077  & 0.708   & 0.546  \\						
			\bottomrule
	\end{tabular}}
	\caption{Performance comparison of using using different types and families of transformer encoder structures in UniST on DHF1K and DAVIS$_{16}$ datasets.}
	\label{table-ablation_diff_encoder}
\end{table}

\subsection{Implementation Details}
To facilitate implementation, the pre-trained MViT-small model\cite{fan2021multiscale} on ImageNet\cite{deng2009imagenet} is employed. The input images of the network are all resized to $224 \times 384$  for training and testing. We follow the experimental settings of prior works  \cite{jain2021vinet, liu2022ds}. For the VSP task, we first pre-train the UniST model on the DHF1K dataset and then fine-tune it on the Hollywood2 and UCF-Sports datasets. And as for the VSOD task, we choose to pre-train the entire model on the image DUTS dataset \cite{wang2017learning}, and then fine-tune it on the  DAVIS$_{16}$ dataset. 




All experimental training process chooses Adam as the optimizer with a learning rate of $1e-4$. The computation platform is configured by two NVIDIA GeForce RTX 4090 GPUs in a distributed fashion, using PyTorch. More implementation details are in the appendix.

\subsection{Evaluation Metrics}
For video saliency detection, we use AUC-Judd $\mathit{AUC}$-$J$, Similarity Metric $\mathit{SIM}$, Linear Correlation Coefficient $\mathit{CC}$, and Normalized Scanpath Saliency $\mathit{NSS}$, following existing work\cite{hu2023tinyhd}.
For video salient object detection, we adopt three evaluation metrics for comparison, including the mean absolute error $\mathit{MAE}$, F-measure $maxF$\cite{achanta2009frequency}, and S-measure $\mathit{S_m}$\cite{fan2017structure}.

\begin{table*}[h]
	\begin{center}
		\centering
		\renewcommand\arraystretch{1.2}
		\resizebox{0.95\linewidth}{!}{
			\begin{tabular}{c | cccc|cccc|cccc}
				\toprule
				\multirow{2}*{\textbf{Method}}  & \multicolumn{4}{c|}{\textbf{DHF1K}} & \multicolumn{4}{c|}{\textbf{Hollywood2}} & \multicolumn{4}{c|}{\textbf{UCF-Sports}}\\
				\cline{2-13}
				&  $\mathit{CC} \uparrow$ & $\mathit{NSS} \uparrow$ & $\mathit{AUC}$-$J$ $\uparrow$  & $\mathit{SIM} \uparrow$&  $\mathit{CC} \uparrow$ & $\mathit{NSS} \uparrow$ & $\mathit{AUC}$-$J$ $\uparrow$    & $\mathit{SIM} \uparrow$ & $\mathit{CC} \uparrow$ & $\mathit{NSS} \uparrow$ & $\mathit{AUC}$-$J$ $\uparrow$   & $\mathit{SIM} \uparrow$ \\
				\Xhline{1pt}
				$\textrm{TASED-Net}_{\mathit{\rm ICCV}^{\prime} 2019}$      & 0.440 & 2.541 & 0.898 & 0.351   & 0.646 & 3.302 & 0.918  & 0.507      & 0.582 & 2.920 & 0.899 & 0.469 \\
				$\textrm{UNIVSAL}_{\mathit{\rm ECCV}^{\prime} 2020}$        & 0.431 & 2.435 & 0.900  & 0.344  & 0.673 & 3.901 & 0.934  & 0.542      & 0.644 & 3.381 & 0.918  & 0.523 \\
				$\textrm{ViNet}_{\mathit{\rm IROS}^{\prime} 2020}$          & 0.460 & 2.557 & 0.900  & 0.352  & 0.693 & 3.730 & 0.930  & 0.550      & 0.673 & 3.620 & 0.924  & 0.522 \\
				$\textrm{VSFT}_{\mathit{\rm TCSVT}^{\prime} 2021}$          & 0.462 & 2.583 & 0.901  & 0.360  & 0.703 & 3.916 & 0.936  & 0.577      & - & - & -  & -   \\
				$\textrm{ECANet}_{\mathit{\rm NeuroComputing}^{\prime} 2022}$      & - & - & -  & -           & 0.673 & 3.380 & 0.929  & 0.526      & 0.636 & 3.189 &  0.917  & 0.498 \\
				$\textrm{STSANet}_{\mathit{\rm TMM}^{\prime} 2022}$         & - & - & -  & -                  & 0.721 & 3.927 & 0.938  & 0.579      & 0.705 & 3.908 &  0.936  & 0.560 \\
				$\textrm{TinyHD-S}_{\mathit{\rm WACV}^{\prime} 2023}$       &  0.492 & 2.873 & 0.907 & 0.388  & 0.690 & 3.815 & 0.935  & 0.561      & 0.624 & 3.280 & 0.918 & 0.510 \\
				$\textrm{TMFI-Net}_{\mathit{\rm TCSVT}^{\prime} 2023}$      & 0.524 & 3.006 & 0.918  & 0.410 & 0.739 & 4.095 & 0.940  & 0.607      & \textbf{0.707} & \textbf{3.863} &  \textbf{0.936}  & 0.565 \\
				\Xhline{1pt}
				\textbf{UniST(\textit{Ours})}   & \textbf{0.541} & \textbf{3.113} & \textbf{0.920}  & \textbf{0.423} & \textbf{0.777}  & \textbf{4.397} & \textbf{0.951}  & \textbf{0.632} & 0.706 & 3.718 & 0.932  &  \textbf{0.576} \\
				\bottomrule
		\end{tabular}}
	\end{center}
	\caption{Comparisons of our method with the other state-of-the-arts on VSP datasets. Our UniST significantly outperforms the previous state-of-the-arts by a large margin.}
	\label{table-sota_vsp}
\end{table*}

\subsection{Ablation Studies}

 To conduct an in-depth analysis of the proposed UniST framework, a range of model baselines and variants are defined as outlined in Table \ref{table-ablation_1}.
 (i) "UniST baseline" denotes a strong saliency model of the proposed UniST framework. It uses MViT-small encoder and task-specific decoders for VSP and VSOD tasks.
 It also combines multi-scale features from the encoder to help boost performance.
 (ii)  "UniST w/SAT" indicates adding the pure saliency-aware transformer upon "UniST baseline". The pure saliency-aware transformer refers to the replacement of the semantic-guided block with a standard 3D convolution, while omitting the saliency transfer operation.
 (iii)  "UniST w/SAT+SGB" indicates adding the proposed semantic-guided block upon "UniST  w/SAT". 
 Similarly, "UniST w/SAT+SGB+ST" denotes the full model after integrating the saliency transfer mechanism.
 
%

To analyze the effectiveness of each part in this work, we investigate the performance of the UniST baseline and its model variants on both DHF1K and DAVIS$_{16}$ datasets, as shown in Table \ref{table-ablation_1}.  It can be observed that the SAT, SGB and ST modules all achieve clear improvement. Specifically, as the core module of the UniST framework, SAT significantly improves the VSP task by 0.011($CC$) on DHF1K, and the VSOD task by 0.011($S_m$) on DAVIS$_{16}$. Finally, the full UniST model achieves remarkable performance gain compared to the UniST baseline.

\subsubsection{Effect of the Number of Sal-Transformer Stages}
There are four stages in the proposed saliency-aware transformer as a default configuration.
Figure \ref{fig_transformer_stages} illustrates the impact of varying the number of sal-transformer stages on task performance across the DHF1K and DAVIS$_{16}$ datasets. Notably, the most optimal performance is attained when the number of sal-transformer stages is set to $\textbf{4}$. These results indicate that UniST needs to gradually fuse the feature of each scale and increase the feature resolution.


\begin{table*}[!htbp]
	\begin{center}
		\centering
		\renewcommand\arraystretch{1.2}
		\resizebox{0.95\linewidth}{!}{
			\begin{tabular}{c|ccc|ccc|ccc|ccc}
				\toprule
				\multirow{2}{*}{\textbf{Method}} & \multicolumn{3}{c|}{\textbf{$\mathit{\textrm{DAVIS}}_{16}$}} & \multicolumn{3}{c|}{\textbf{FBMS}} & \multicolumn{3}{c|}{\textbf{ViSal}} & \multicolumn{3}{c}{\textbf{SegV2}}       \\                     
				\cline{2-13} 
				& $\mathit{MAE}\downarrow$ & $S_m\uparrow$ & $maxF\uparrow$ & $\mathit{MAE}\downarrow$ & $S_m\uparrow$ & $maxF\uparrow$ & $\mathit{MAE}\downarrow$ & $S_m\uparrow$ & $maxF\uparrow$ & $\mathit{MAE}\downarrow$ & $S_m\uparrow$ & $maxF\uparrow$ \\ 
				\Xhline{1pt}
				$\textrm{SSAV}_{\mathit{\rm CVPR}^{\prime}2019}$          & 0.028       & 0.893          & 0.861         & 0.040           & 0.879         & 0.865          & 0.020      & 0.943        & 0.939         & 0.023      & 0.851          & 0.801 \\
				$\textrm{CAS}_{\mathit{\rm TNNLS}^{\prime}2020}$          & 0.032       & 0.873          & 0.860         & 0.056          & 0.856          & 0.863          & -          & -             & -          & 0.029        & 0.820         & 0.847 \\
				$\textrm{PCSA}_{\mathit{\rm AAAI}^{\prime}2020}$          & 0.022       & 0.902          & 0.880         & 0.040           & 0.868         & 0.837         & 0.017       & 0.946          & 0.940      & 0.025       & 0.865       & 0.810 \\
				$\textrm{FSNet}_{\mathit{\rm ICCV}^{\prime}2021}$         & 0.020       & \textbf{0.920} & \textbf{0.907} & 0.041         & 0.890          & 0.888         & -           & -              & -          & 0.023       & 0.870       & 0.772 \\
				$\textrm{ReuseVOS}_{\mathit{\rm CVPR}^{\prime}2021}$      & 0.019       & 0.883         & 0.865          & 0.027          & 0.888          & 0.884          & 0.020      & 0.928          & 0.933       & 0.025       & 0.844         & 0.832 \\
				$\textrm{TransVOS}_{\mathit{\rm PrePrint}^{\prime}2021}$  & 0.018       & 0.885         & 0.869          & 0.038          & 0.867          & 0.886         & 0.021       & 0.917           & 0.928       & 0.024       & 0.816        & 0.800 \\ 
				$\textrm{UFO}_{\mathit{\rm TMM}^{\prime}2023}$            & 0.036       & 0.864         & 0.828          & 0.028          & 0.894          & \textbf{0.890} & 0.011      & \textbf{0.953} & 0.940      & 0.022 & 0.892    & \textbf{0.863} \\ 
				\Xhline{1pt}
				\textbf{UniST(\textit{Ours})}                             & \textbf{0.018} & 0.904      & 0.873          & \textbf{0.027} & \textbf{0.902} & 0.884        & \textbf{0.011} & 0.952       & \textbf{0.952} & \textbf{0.017} & \textbf{0.897} &  0.854    \\
				\bottomrule
			\end{tabular}
		}
	\end{center}
	
	\caption{Comparisons of our method with the other state-of-the-arts on VSOD datasets. Our UniST outperforms the previous state-of-the-arts on most of the metrics on these four datasets.}
	\label{table-sota_vsod}
\end{table*}

\subsubsection{Image Encoder vs. Video Encoder}
A further performance comparison is conducted using image and video encoders from two transformer families.
As shown in Table \ref{table-ablation_diff_encoder}, there is minimal variation in the model's performance when employing either the image or video encoder on the DHF1K dataset.
Conversely, on the DAVIS$_{16}$ dataset, the model employing the image encoder surpasses its video encoder-based counterpart by a significant margin. Due to the limited size of the DAVIS$_{16}$ dataset, it is difficult to converge the model to an optimal state by training on this dataset directly.
Therefore, the image MViT model  with the best performance on both datasets is chosen as the default encoder in our work.



%

\begin{figure}[t]
	\centering
	\includegraphics[scale=0.35]{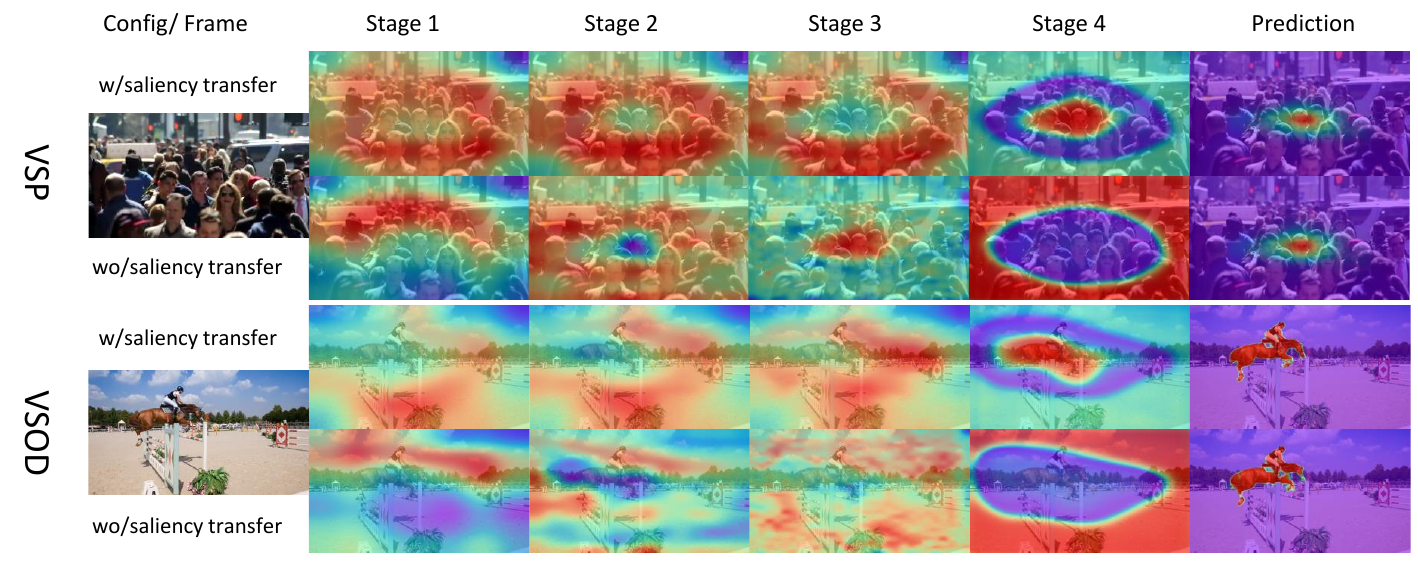}
	\caption{Comparing the visualization results of attention scores in different sal-transformer stages with and without the saliency transfer mechanism.
	}
	\label{fig-saliency_transfer}
\end{figure}

\subsubsection{Further Analysis of Saliency Transfer} 

Figure \ref{fig-saliency_transfer} shows the visualized results of the attention scores in various sal-transformer stages with/without the saliency transfer mechanism. Compared to configurations without saliency transfer,  the gradual integration of the attention score significantly helps the model learn a more discriminative feature, thus resulting in better visualization results.

\subsection{Comparision with State-of-the-arts}

\begin{figure*}[!htbp]
	\centering
	\includegraphics[width=\textwidth]{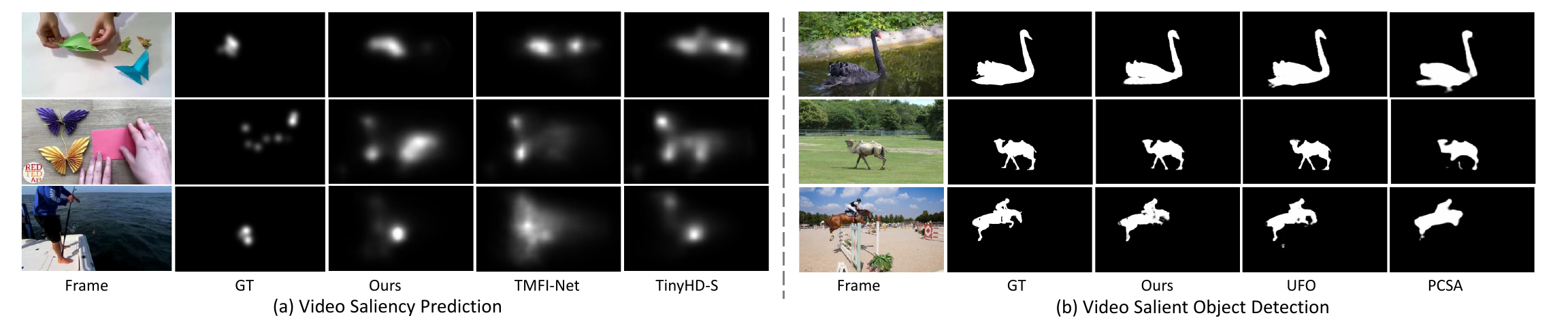}
	
	\caption{Qualitative results of our method compared with other state-of-the-art methods on VSP and VSOD tasks. GT means the ground-truth.
	}
	\label{fig-vis_results}
	
\end{figure*}

\subsubsection{Video Saliency Prediction}
The proposed UniST is compared with recent state-of-the-art works on three video saliency datasets as shown in Table \ref{table-sota_vsp}. Experimental results in the table highlight the superiority of the proposed unified scheme, as it outperforms the other comparable  works on almost all datasets and metrics. Notably, UniST  significantly surpasses the previous top-performing methods, such as TMFI-Net \cite{zhou2023transformer} and VSFT \cite{ma2022video}. Compared to TMFI-Net, UniST improves the average $CC$ and $SIM$ performance on the DHF1K and Hollywood2 datasets by 4.2\% and 3.1\%, respectively, and becomes the new state-of-the-art on both benchmarks. Moreover, on the UCF-Sports dataset, our method demonstrates competitive performance comparable to TMFI-Net.  Figure \ref{fig-vis_results}(a) shows that the prediction results of the proposed work are more likely as the ground-truth in comparison to other works. 
These significant improvements indicate that our saliency-aware transformer is well suited for the spatio-temporal modeling required in video saliency prediction.

\subsubsection{Video Saliency Object Detection} 
Table \ref{table-sota_vsod} shows a comparison of the proposed UniST method against existing state-of-the-art, including UFO \cite{su2023unified}, TransVOS \cite{mei2021transvos} and ReuseVOS \cite{park2021learning}, on four video datasets.  The UniST is able to achieve the best performance  across a majority of metrics for all four datasets. As opposed to the UFO model, which introduces the additional COCO-SEG dataset\cite{lin2014microsoft}, the UniST simply uses the image saliency dataset DUTS \cite{wang2017learning} for pre-training to achieve comparable performance. 
Compared to TransVOS \cite{mei2021transvos}, the UniST achieves $2.1\%$ and $4\%$ $S_m$ improvement on the two major datasets, DAVIS$_{16}$ and FBMS, respectively. 
These results clearly show that our UniST can successfully utilize a generalized framework to solve the video salient object detection problem.
Figure \ref{fig-vis_results}(b) shows some qualitative results which indicate that our method is closer to the ground-truth. More visualization results can be found in the supplementary.

\section{Conclusion}

In this paper, we propose a novel unified saliency transformer framework, UniST, to unify the modeling paradigms for video saliency prediction and video saliency object detection tasks. 
By capturing the spatial features of video frame sequences through a visual feature encoder, the subsequent  saliency-aware transformer not only helps the model to capture the spatio-temporal information in the sequence of image features but also progressively increases the scale of the features for saliency prediction. 
Finally, a task-specific decoder is devised to leverage the transformer's output features, and facilitate the output predictions for each individual task.
Extensive experiments demonstrated the effectiveness of the proposed method and also showed its significantly better performance on seven popular benchmarks compared to the previous state-of-the-art methods.

\subsubsection{Limitations} Although our model has good performance in saliency prediction, the improvements on detection are not such significant. We conjecture that this is due to the lack of temporal information in the image dataset used for VSOD pre-training, making it difficult for the saliency-aware transformer to provide spatio-temporal modeling. We believe that the UniST can be further improved after adding more video datasets. 

\subsubsection{Acknowledgments}
This work is supported by the National Natural Science Foundation
of China (No.61971352, No.62271239), Ningbo Natural Science
Foundation (No.2021J048, No.2021J049), Jiangxi Double Thousand
Plan (No.JXSQ2023201022), Fundamental Research Funds for the
Central Universities (No.D5000220190), Innovative Research Foundation of Ship General Performance (No.25522108).

\appendix

\bibliography{yourbibfile}

\appendix

\newpage

\section{More Implementation Details} 


To facilitate implementation, the pre-trained MViT-small model\cite{fan2021multiscale} on ImageNet\cite{deng2009imagenet} is employed. The input images of the network are all resized to $224 \times 384$  for training and testing.
\textbf{(1):} On video saliency detection, we set batch size to 6 and set $T$ to 16 and use the annotations in the middle of the $T$ frames for supervision. Like prior works \cite{jain2021vinet, ma2022video}, to evaluate the Hollywood2 and UCF-Sports benchmarks, the whole model needs to be pre-trained on the DHF1K dataset first, and then fine-tuned on the Hollywood2 and UCF-Sports training sets. 
\textbf{(2):}  On video salient object detection, we set batch size to 16 and set $T$ to 5. We follow the common practice for most methods\cite{zhang2021dynamic, liu2022ds}, which first pre-trained their models on the static saliency dataset, such as DUTS\cite{wang2017learning}.  And we then proceed to fine-tune the UniST model on the DAVIS$_{16}$ training set. 
All experimental training process chooses Adam as the optimizer with a learning rate of $1e-4$. The computation platform is configured by two NVIDIA GeForce RTX 4090 GPUs in a distributed fashion, using PyTorch.

\subsection{Saliency-aware Transformer}
There are four sal-transformer stages in saliency-aware transformer. Each stage has one sal-transformer block, where the number of multi-heads is set to 2 by default. The convolution kernel $K$ and stride size $S$ of the $Conv3dLN_Q$ operation are constantly 3 and 1, respectively, while both are of size $\{ 2, 4, 8, 16 \}$ for the $Conv3dLN_K$ and $Conv3dLN_V$ operations, respectively, see table below for specific configuration parameters. 

\begin{table}[!htbp]
	\resizebox{\linewidth}{!}{
		\begin{tabular}{l|l|l|l}
			& $Conv3dLN_Q$                                             & $Conv3dLN_K$                                            & $Conv3dLN_V$                                             \\ \hline
			\multicolumn{1}{l|}{Stage1} & \begin{tabular}[c]{@{}l@{}} $ K= 3\times 3 \times 3$ \\ $ S= 1\times 1 \times 1$ \\  $ P= 1\times 1 \times 1$ \end{tabular} & \begin{tabular}[c]{@{}l@{}}$ K=2\times 2 \times 2$\\ $S=2\times 2 \times 2$\\ $P=0\times 0 \times 0$\end{tabular} & \begin{tabular}[c]{@{}l@{}} $K=2\times 2 \times 2$\\ $S=2\times 2 \times 2$\\ $P=0\times 0 \times 0$\end{tabular} \\ \hline
			\multicolumn{1}{l|}{Stage2} & \begin{tabular}[c]{@{}l@{}}$K= 3\times 3 \times 3$ \\  $S= 1\times 1 \times 1$ \\  $P=1\times 1 \times 1$\end{tabular} & \begin{tabular}[c]{@{}l@{}} $K= 4\times 4 \times 4$\\ $S=4\times 4 \times 4$\\  $P=0\times 0 \times 0$\end{tabular} & \begin{tabular}[c]{@{}l@{}}$K=4\times 4 \times 4$\\  $S=4\times 4 \times 4$\\  $P=0\times 0 \times 0$\end{tabular} \\ \hline
			\multicolumn{1}{l|}{Stage3} & \begin{tabular}[c]{@{}l@{}} $K=3\times 3 \times 3$ \\  $S= 1\times 1 \times 1$ \\  $P=1\times 1 \times 1$\end{tabular} & \begin{tabular}[c]{@{}l@{}}  $K=8\times 8 \times 8$\\ $S= 8\times 8 \times 8$\\  $P=0\times 0 \times 0$\end{tabular} & \begin{tabular}[c]{@{}l@{}}  $K=8\times 8 \times 8$\\  $S=8\times 8 \times 8$\\  $P=0\times 0 \times 0$\end{tabular} \\ \hline
			\multicolumn{1}{l|}{Stage4} & \begin{tabular}[c]{@{}l@{}} $K=3\times 3 \times 3$ \\  $S= 1\times 1 \times 1$ \\  $P=1\times 1 \times 1$\end{tabular} & \begin{tabular}[c]{@{}l@{}}  $K=16\times 16 \times 16$\\  $S=16\times 16 \times 16$\\  $P=0\times 0 \times 0$\end{tabular} & \begin{tabular}[c]{@{}l@{}} $K=16\times 16 \times 16$\\  $S=16\times 16 \times 16$\\  $P=0\times 0 \times 0$\end{tabular} \\ 
	\end{tabular}}
	\label{tab_convln_config}
\end{table}

\begin{figure*}[!htbp]
	\centering
	\includegraphics[width=\textwidth]{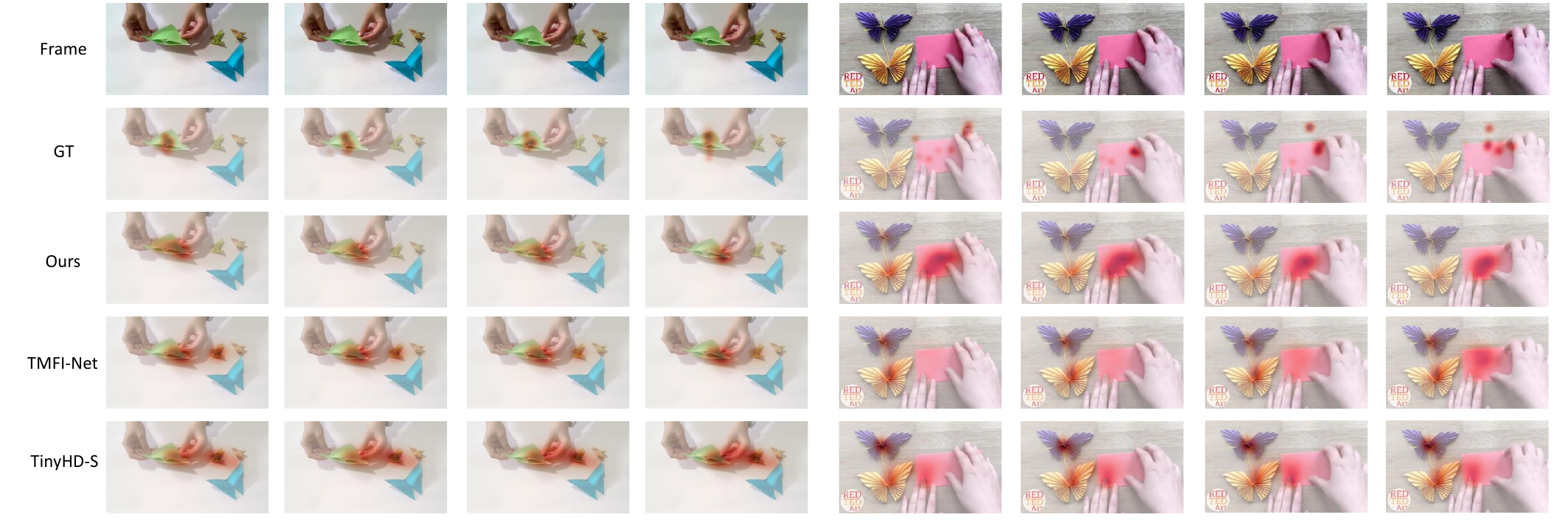}
	
	\caption{Qualitative results of our method compared with other state-of-the-art methods on Video Saliency Prediction task.
	}
	\label{fig-vsp_results}
\end{figure*}

\begin{figure*}[h]
	\centering
	\includegraphics[width=\textwidth]{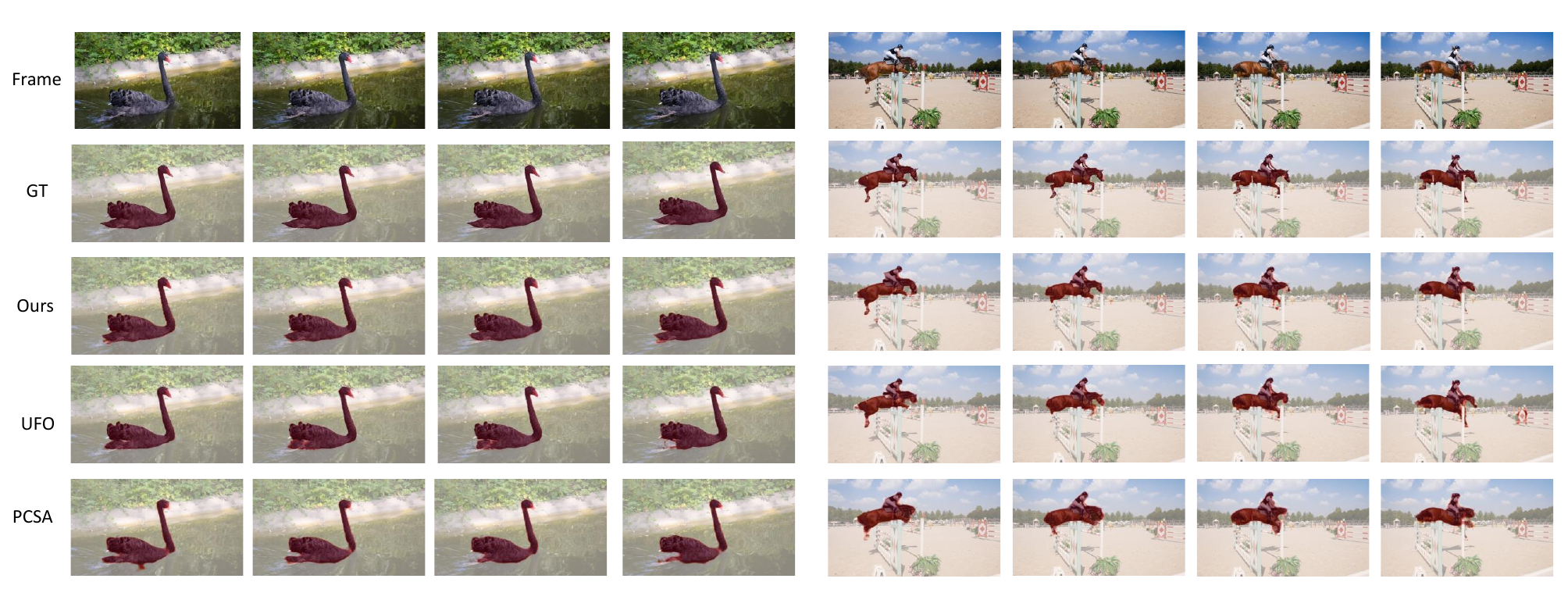}
	
	\caption{Qualitative results of our method compared with other state-of-the-art methods on Video Salient Object Detection task.
	}
	\label{fig-vsod_results}
\end{figure*}

\subsection{Training Objectives of Video Saliency Detection}
We refer to the training paradigm of multiple loss functions in \cite{tsiami2020stavis, chang2021temporal}, which contains: Kullback-Leibler ($KL$) divergence, Linear Correlation Coefficient ($CC$) and Similarity Metric ($SIM$). Assuming that the predicted saliency map is $P_{VSP} \in [0, 1]$, the labeled binary fixation map is $G_{fix} \in \{0, 1\}$, and the dense saliency map generated by the fixation map is $G_{den} \in [0,1]$, then $L_{KL}$, $L_{CC}$, and $L_{NSS}$ are employed to signify three different loss functions, respectively. The first is the $KL$ loss between the predicted map $P_{VSP}$ and the dense map $G_{den}$:

\begin{equation}
	L_{KL} (P_{VSP}, S_{den}) = \sum_{x}^{}G_{den}(x)ln\frac{G_{den}(x)}{P_{VSP}(x)}
\end{equation}

\noindent where $x$ represents the spatial domain of a saliency map. The second loss function is based on the $CC$ that has been widely used in saliency evaluation, and used to measure the linear relationship between the predicted saliency map $P_{VSP}$ and the dense map $G_{den}$:

\begin{equation}
	L_{CC} (P_{VSP}, G_{den}) = - \frac{cov(P_{VSP}, G_{den})}{\rho(P_{VSP})\rho(G_{den})}
\end{equation}

\noindent where $cov(\cdot)$ and $\rho(\cdot)$ represent the covariance and the standard deviation respectively. The last one is derived from the $SIM$, which can measure the similarity between two distributions:
\vspace{-10pt}
\begin{equation}
	\begin{aligned}
		L_{SIM}(P_{VSP}, G_{den}) =  \sum_{x}^{} min\{\zeta(P_{VSP}(x)), \zeta(G_{den}(x)) \} \\
	\end{aligned}
\end{equation}

\noindent where $\zeta$ represents the normalization operation. The weighted summation of the above $KL$, $CC$ and $SIM$ is taken to represent the final loss function:
\vspace{-5pt}
\begin{equation}
	L_{VSP} =  L_{KL} + \lambda_1 L_{CC} + \lambda_2 L_{SIM}
\end{equation}

\noindent where $\lambda_1$, $\lambda_2$ are the weights of $CC$ and $SIM$, respectively. We set $\lambda_1 = \lambda_2 = -0.1$ in our implementation.  

\subsection{Training Objectives of Video Salient Object Detection}
We refer to the training paradigm in \cite{gu2020pyramid} that uses the binary cross entropy loss function. We denote our VSOD prediction as $P_{VSOD}$, and the Ground Truth of the saliency map is $G_{VSOD}$. Then binary cross entropy loss $L_{VSOD}$ can be defined as

\begin{equation}
	\begin{aligned}
		L_{VSOD} (P_{VSOD}, G_{VSOD}) =& - \frac{1}{N} \sum_{i=1}^{N} [g_i log(p_i) \\
		&+ (1-g_i) log(1-p_i)]
	\end{aligned}
\end{equation}

\noindent where $N$ is the number of pixels.

\section{More Qualitative Results}
In Figures \ref{fig-vsp_results} and \ref{fig-vsod_results}, we show the prediction results of the proposed UniST model, as well as the predictions of other SOTA methods. The UniST produces significantly better results than the other methods, being closer to the ground-truth on both tasks.

\end{document}